\documentclass{article}

\usepackage{PRIMEarxiv}

\usepackage[utf8]{inputenc} % allow utf-8 input
\usepackage[T1]{fontenc}    % use 8-bit T1 fonts
\usepackage{hyperref}       % hyperlinks
\usepackage{url}            % simple URL typesetting
\usepackage{booktabs}       % professional-quality tables
\usepackage{amsfonts}       % blackboard math symbols
\usepackage{nicefrac}       % compact symbols for 1/2, etc.
\usepackage{microtype}      % microtypography
\usepackage{lipsum}
\usepackage{fancyhdr}       % header
\usepackage{graphicx}       % graphics
\graphicspath{{media/}}     % organize your images and other figures under media/ folder

\usepackage[textsize=tiny]{todonotes}
\usepackage{amsmath}
\usepackage{amssymb}
\usepackage{cleveref}

\newcommand{\ouracronym}{{\sc unify}}

\DeclareMathOperator*{\argmin}{arg\,min}

\title{UNIFY: a Unified Policy Designing Framework for Solving Constrained Optimization Problems with Machine Learning}

\author{
  Mattia Silvestri\thanks{Equal contributors.},
    Allegra De Filippo\footnotemark[1],
    Michele Lombardi\footnotemark[1],
    Michela Milano \\
  Dipartimento di Informatica - Scienza e Ingegneria \\
  University of Bologna\\
  \texttt{\{mattia.silvestri4,allegra.defilippo,michele.lombardi2,michela.milano\}@unibo.it} \\
}

\begin{document}
\maketitle

\begin{abstract}

The interplay between Machine Learning (ML) and Constrained Optimization (CO) has recently been the subject of increasing interest, leading to a new and prolific research area covering (e.g.) Decision Focused Learning and Constrained Reinforcement Learning.
Such approaches strive to tackle complex decision problems under uncertainty over multiple stages, involving both explicit (cost function, constraints) and implicit knowledge (from data), and possibly subject to execution time restrictions. 
While a good degree of success has been achieved, the existing methods still have limitations in terms of both applicability and effectiveness.
For problems in this class, we propose \ouracronym{}, a unified framework to design a solution policy for complex decision-making problems.
Our approach relies on a clever decomposition of the policy in two stages, namely an unconstrained ML model and a CO problem, to take advantage of the strength of each approach while compensating for its weaknesses.
With a little design effort, \ouracronym{} can generalize several existing approaches, thus extending their applicability.
We demonstrate the method effectiveness on two practical problems, namely an Energy Management System and the Set Multi-cover with stochastic coverage requirements.
Finally, we highlight some current challenges of our method and future research directions that can benefit from the cross-fertilization of the two fields.

\end{abstract}

% INTRODUCTION
\section{Introduction}%
\label{sec:introduction}

Methods for combining Machine Learning (ML) and Constrained Optimization (CO) for decision support have attracted considerable interest in recent years.
This is motivated by the possibility to tackle complex decision making problems subject to uncertainty (sometimes over multiple stages), and having a partially specified structure where knowledge is available both in explicit form (cost function, constraints) and implicit form (historical data or simulators).

%This is motivated on one side by a growing recognition that the two fields have a strong connection, on the other by difficulties faced by pure approaches of either class on some practical problems.

%For example, on one hand, cost and constraint parameters of optimization models have been estimated via statistics or ML for decades, but little attention was paid to this interaction. On the other hand, pure data-driven methods such as Reinforcement Learning (RL) have trouble dealing with hard constraints and combinatorial decision spaces, while pure CO methods often struggle with robustness.

%Motivated by the opportunity to obtain improvements by combining ML and CO, multiple lines of research have emerged, such as Decision Focused Learning, Constrained Reinforcement Learning, or Algorithm Configuration.

%These methods usually rely on some available information (e.g. observable variables, past solutions) and also on the declarative formulation of the problem, namely the objective function and the constraints. 

As a practical example, an Energy Management Systems (EMS) needs to allocate minimum-cost power flows from different Distributed Energy Resources (DERs) \cite{6162768}. Based on actual energy prices, and forecasts on the availability of DERs and on consumption, the EMS decides which power generators should be used and whether the surplus should be stored or sold to the market.
Such a problem involves hard constraints (maintaining power balance, power flow limits), a clear cost structure, elements of uncertainty that are partially known via historical data, and multiple decision stages likely subject to execution time restrictions.
In this type of use case, pure CO methods struggle with robustness and scalability, while pure ML methods such as Reinforcement Learning (RL) have trouble dealing with hard constraints and combinatorial decision spaces.
Motivated by the opportunity to obtain improvements via a combination of ML and CO, multiple lines of research have emerged, such as Decision Focused Learning, Constrained Reinforcement Learning, or Algorithm Configuration.
While existing methods have obtained a good measure of success, to the best of the authors knowledge no existing method can deal with all the challenges we have identified.

Ideally, one wishes to obtain \emph{a solution policy} capable of providing feasible (and high-quality) solutions, handling robustness, taking advantage of existing data, and with a reasonable computational load.
In this paper, we argue this can be achieved by introducing \emph{a unification framework for a family of existing ML \& CO approaches}, in particular: 1) Decision Focused Learning, 2) Constrained Reinforcement Learning, 3) Algorithm Configuration and 4) Stochastic Optimization algorithms.
We assume to have access to problem knowledge in the form of both a declarative formulation (namely an objective function and a set of constraints), and of historical data.
The framework is then based on a two-step policy decomposition, respectively into an unconstrained ML model and a CO problem.
The interface between the two components consists of a new set of \emph{``virtual'' model parameters}, which can serve as an additional (and potentially very useful) design handle.
Since the approach is decomposition-based, multiple learning and optimization methods can be used for its implementation. In our presentation we use RL for the learning task, due to its ability to handle both non-differentiable loss functions and sequential problems.
We refer to our method as \ouracronym{}.

The paper is structured as follows: we describe two motivating use cases in \cref{sec:usecases}; in \Cref{sec:formalization} we formalize the approach, while in \Cref{sec:generalization} we show its relation with other hybrid and traditional methods for decision support. We present an empirical evaluation on the two use cases in \Cref{sec:experiments}, while concluding remarks are in \Cref{sec:conclusion}.

% USE CASES DESCRIPTION
\section{Use Cases Description}
\label{sec:usecases}

We rely on two use cases to motivate and demonstrate our approach, i.e. an Energy Management System and a simplified production scheduling problem, for which both \textit{implicit and explicit knowledge} is available in the form of:
    \begin{itemize}
        \item historical data or simulators;
        \item an objective function;
        \item problem constraints.
    \end{itemize}
%

%In this section we describe the CO problems that we will use to introduce, explain and experimentally demonstrate the \ouracronym{} framework.

% ENERGY MANAGEMENT SYSTEM
\paragraph{Energy Management System}

First, we consider a real-world EMS that requires allocating minimum-cost power flows from different DERs \cite{6162768}. The problem has a high level of uncertainty, which stems from uncontrollable deviations from the planned consumption load and the presence of Renewable Energy Sources. 
Based on actual energy prices, and forecasts on the availability of DERs and on consumption, the EMS decides: 1) how much energy should be produced; 2) which generators should be used; 3) whether the surplus of energy should be stored or sold to the energy market.
The problem admits both a single-stage and a sequential formulation; in the former, a plan for 96 time units (each one 15 minutes long) must be prepared one day in advance; in the latter, routing decisions are made each time unit for the next one, until the end of horizon is reached.

In this case, historical costs, forecasts, and actual power generation and consumption represent the available data. The objective function is the power flow cost and the constraints impose power balance and power flow limits.
This problem was tackled in \cite{de2020blind} by introducing a virtual model parameter. In particular, it was shown how associating a (normally absent) cost to the storage equipment can improve the performance of a baseline optimization method.
An effective, dedicated, algorithm configuration approach was then developed.

A complete model is provided in the supplemental material.

% This problem admits both a single-stage and a sequential formulation, and complete model is available in Appendix~XYZ. The model features continuous variables, linear inequality and equality constraints, and a linear cost.

%The former was employed by \cite{de2020blind}, which introduced (albeit in a preliminary form) some of the key ideas that are generalized in this paper.

% SET MULTI-COVER
\paragraph{Set Multi-cover}

Second, we consider a simplified production planning problem, modeled as a Set Multi-cover Problem with stochastic coverage requirements \cite{hua2009exact}. Given a universe $N$ containing $n$ elements and a collection of sets over $N$, the Set Multi-cover Problem requires finding a minimum size sub-collection of the sets such that coverage requirements for each element are satisfied. The sets may represent products that need to be manufactured together, while the coverage requirements represent product demands.

We consider a version of the problem where sets have non-uniform manufacturing costs, and where the demands are stochastic and unknown at production time. Unmet demands can then be satisfied by buying additional items, but at a higher cost.
The requirements are generated according to a Poisson distribution, and we assume the existence of a linear relationships between an observable variable $o \in \mathbb{R}$ and the Poisson rates $\lambda$ for each product, i.e. $\lambda_j = a_j o \quad \forall j \in N $.

In this case, the historical data is represented by a dataset $\{o_{i}, d_{i}\}_{i=1, \dots, m}$, where $d$ are the demands and $m$ is the dataset size. The objective is to minimize the cost of manufacturing and buying additional items; the constraints require to manufacture products in the same set together.

The full problem description is available in the supplementary material.

% UNIFY
\section{The \ouracronym{} Framework}%
\label{sec:formalization}

% Given the available information and the declarative formulation of the CO problem, defined by an objective function $f$ and a set of constraints $C$, our objective is to learn a policy that provides a high-quality solution to the problem itself. The term ``policy'' is a reference to Reinforcement Learning, which we will eventually employ; for sake of simplicity, however, we will first formalize the approach for single-stage problems, where all decisions must be taken at once.

Our objective is to learn a policy that provides a high-quality solution to the problem itself. The term ``policy'' is a reference to Reinforcement Learning, which we will eventually employ; for sake of simplicity, however, we will first formalize the approach for single-stage problems, where all decisions must be taken at once.

Formally, let $x \in X$ be a vector representing observable information and let $z$ be a vector representing the decisions, which must come from the feasible set $C(x)$.
In the case of the EMS, $x$ refers to the production and consumption forecasts and $z$ are the power flow values. 
For the Set Multi-cover, the observable $x$ corresponds to $o$, whereas the decision variables $z$ are the amount of sets to manufacture.

Typically, $z$ will have a fixed size and $C(x)$ will specify which vectors in the decision space are feasible. In general, however, even the size of $z$ may depend on the observable (e.g. decide which transplants $z$ to perform, given a pool $x$ of patients and donors).
In all cases, we aim at defining a function $\pi(x, \theta)$, with parameter vector $\theta \in \Theta$, such that:
\begin{equation}
    \pi : (x, \theta) \mapsto z \in C(x)
\end{equation}
The $\pi$ function is our \emph{constrained} policy. In both the use cases, $\pi$ should provide feasible decisions, i.e. power flow values or units of each set to be manufactured.
We wish to choose $\theta$ to minimize a cost metric on a target distribution. The corresponding training problem can be formalized as:
\begin{align}
    \argmin_{\theta \in \Theta} \mathbb{E}_{(x_+, x) \sim p} \left[ f(x_+, x, \pi(x, \theta)) \right]
    \label{eqn:raw_training}
\end{align}
where $f(x_+, x, z)$ is a function returning the cost of decisions $z$, when $x$ is observed and uncertainty unfolds with outcome $x_+$. We assume  without loss of generality that $x_+ \in X$; the term $p$ refers to the training distribution (e.g. approximated via a simulator or a training set). 

Given a solution $\theta^*$ to the training problem, we can then perform inference (i.e. solve the decision problem) by observing $x$ and evaluating $\pi(x, \theta^*)$. Unfortunately, \Cref{eqn:raw_training} is hard to solve, 
since $\pi$ will typically be \emph{trained on a large dataset} and it is expected to \emph{always return feasible decisions} from a space \emph{possibly lacking a fixed structure}.

Our key insight is that \Cref{eqn:raw_training} can be made easier to handle by decomposing the policy into a traditional Machine Learning model and a traditional Constrained Optimization problem. Formally, the policy can be reformulated as:
\begin{equation}
    \pi(x, \theta) = g(x, h(x, \theta))
    \label{eqn:om_pi}
\end{equation}
with:
\begin{equation}
    g(x, y) \equiv \argmin_{z \in \tilde{C}(x, y)} \tilde{f}(x, y, z)
    \label{eqn:om_g}
\end{equation}
where $h(x, \theta)$ is the ML model and $g(x, y)$ is a function defined via the constrained optimization problem.
The term $y$, which corresponds to the output of the ML model, is referred to as \emph{virtual parameter vector}.  Its value may have an impact both on the feasible set $\tilde{C}$ and on the cost $\tilde{f}$ of the optimization problem, which are referred to as \emph{virtual feasible set} and \emph{virtual cost}.
For example, in the EMS, $y$ may correspond to the (normally absent) cost associated to the storage system mentioned in \cref{sec:usecases}, and as a side effect require to use a modified cost function $\tilde{f}$. It is also possible to use an existing set of parameters as $y$: for instance, in the Set Multi-cover use case, the ML model can be designed to predict the coverage requirements $d$, thus requiring no modification in the structure of the feasible set.

% PROPERTIES OF THE APPROACH
\paragraph{Properties of the Approach}

There are a few things to notice about \Cref{eqn:om_pi} and~\eqref{eqn:om_g}.
First, in the reformulated policy it is comparatively easy to enforce feasibility and to deal with combinatorial decision spaces: since this is done \emph{outside} of the ML model, any traditional constrained optimization method can be used (e.g. Mathematical Programming, Constraint Programming, or the Alternating Direction Method of Multipliers).

Second, the reformulated training problem is:
\begin{equation}
\label{eqn:om_training}
\begin{aligned}
\argmin_{\theta \in \Theta}\ & \mathbb{E}_{(x_+, x) \sim p} \left[ f(x_+, x, z) \right] \\
			& z = g(x, y) = \argmin_{z \in \tilde{C}(x, y)} \tilde{f}(x, y, z) \\
			& y = h(x, \theta)
\end{aligned}
\end{equation}
This is a bilevel problem, combining unconstrained optimization on $\theta$ and constrained optimization on $z$.
Several techniques are available to tackle \Cref{eqn:om_training}. For example, if $h$ is differentiable, one may use the classical subgradient method to reach a local optimum: in the ``forward pass'' we evaluate $h$ and compute $z$, in the ``backward pass'' we fix the value of $z$ and differentiate $h$ w.r.t. the parameters $\theta$. 

Third, since most bilevel optimization methods rely on separate steps to deal with $\theta$ and $z$, the optimization problem that defines $g$ can be built on the fly, adjusting the size of the decision vector $z$ as needed (e.g. scaling the number of possible transplants with the patient/donor pool).

Finally, the reformulated policy features a vector (the virtual parameters $y$) that is absent in the original formulation. Choosing $y$ is a design decision, and the trickiest step for applying our method.
A conservative approach consists in: 1) starting from a mathematical model of the original decision problem (e.g. routing, assignment, selection); 2) selecting a set of parameters from the model to be used as $y$.

While this approach is viable, it misses a key opportunity: in fact, as the notation and terminology for $\tilde{C}$ and $\tilde{f}$ imply, the feasible set and cost function of the inner optimization problem $g$ may \emph{differ from the original ones}.
For example, it is possible to introduce penalties or rewards, buffers on constraint satisfaction thresholds, or even to relax constraints or to remove cost terms.
This element of flexibility provides a new design handle that can be used to ``partition'' the challenging aspects of the original problem into either $h$ or $g$, depending on which of the two components is best suited to deal with them. While this is a powerful new mechanism to tackle complex problems, it does require a degree of creativity and expertise to be effectively used. We provide a few examples of how to do it in \Cref{sec:experiments}.

% SEQUENTIAL FORMULATION AND RL
\paragraph{Sequential Formulation and RL}

The training formulation from \Cref{eqn:om_training} applies to single-stage problems, but it can be extended to sequential decision-making. Formally, we can view the problem as a Markov Decision Process $\langle X, Z, p_+, f, p_1, \gamma \rangle$ where $X$ is the set of possible (observable) states and $Z$ is the decision space, $p_+$ is the probability distribution for state transitions, $f$ is the cost function, $p_1$ is the probability distribution for the initial state, and $\gamma \in (0, 1]$ is a discount factor.
At each step of the sequential process we will have access to a distinct observed state $x_k$ and we will output a distinct decision vector $z_k$; the value $p_+(x_{k+1}, x_k, z_k)$ denotes the probability of reaching state $x_{k+1}$ when applying decisions $z_k$ on state $x_k$; the cost function $f$ is the same as in \Cref{eqn:raw_training} and~\eqref{eqn:om_training}. Within this framework, the training problem becomes:
\begin{equation}
\label{eqn:om_seq_training}
\begin{aligned}
\argmin_{\theta \in \Theta}\ & \mathbb{E}_{\tau \sim p} \left[ \sum_{k=1}^{eoh} \gamma^k f(x_{k+1}, x_k, z_k) \right] \\
			& z_k = g(x_k, y_k) = \argmin_{z \in \tilde{C}(x_k, y_k)} \tilde{f}(x_k, y_k, z_k) \\
			& y_k = h(x_k, \theta)
\end{aligned}
\end{equation}
where $y_k$ is the ML model output for step $k$, and $\tau$ refers to a trajectory, i.e. to a sequence of states, ML outputs, and decisions $\{(x_k, y_k, z_k)\}_{k=1}^{eoh}$. The probability of a trajectory is given by $p(\tau) = p_1(x_1) \prod_{k = 1}^{eoh} p_+(x_{k+1}, x_k, z_k)$. The term $eoh$ is the End Of Horizon, and it can be infinite if the sequence is not upper-bounded. In this case, the discount factor $\gamma$ should be strictly lower than 1, while $\gamma$ should be equal to 1 for decision problems over a finite horizon.

\Cref{eqn:om_seq_training} can be interpreted as defining a Reinforcement Learning problem, where the $h$ plays the role of a conventional RL policy, and $g$ can be viewed at training time as part of the environment. This is an important observation since it implies that \emph{any RL approach can be used}.
Once an optimized parameter vector $\theta^*$ has been obtained, the reformulated policy is given by $g(x, h(x, \theta^*))$ as usual.
The sequential formulation directly generalizes the single step one, meaning that RL algorithms can be used to tackle \Cref{eqn:om_training}. In fact, this is exactly what we do in our experimentation since it simplifies the exposition.

% RELATION TO OTHER APPROACHES
\section{Relation to Other Approaches}%
\label{sec:generalization}

% DECISION FOCUSED LEARNING
\paragraph{Decision Focused Learning (DFL)}

Approaches in this class seek to train estimators for parameters of optimization models, while explicitly \emph{accounting for the impact that estimation inaccuracy has on the decisions}.
The idea gained attention after the seminal work by \cite{donti2017task}, with several new methods being proposed -- and well covered in a recent survey by \cite{DBLP:conf/ijcai/KotaryFHW21}.

While the original approach was limited to convex costs and constraints, subsequent works have tackled combinatorial problems, either in an approximate fashion via continuous relaxations \cite{wilder2019melding}, or in an exact fashion by assuming a fixed feasible space and linear costs \cite{elmachtoub2017smart,vlastelica2019differentiation}. Outer and inner relaxations have also been used to improve scalability \cite{DBLP:conf/nips/MandiG20,DBLP:conf/ijcai/MulambaMD0BG21}.

DFL can be interpreted as solving (via subgradient optimization) a close variant of \Cref{eqn:om_training}, with some restrictions and one generalization. In particular, all DFL approaches lack virtual parameters: the ML model is instead expected to estimate part of the future state, i.e. $y$ is a prediction for a portion of $x_+$. For example, the estimator might look at current traffic to predict future travel times.
Additionally, $\tilde{f}$ always corresponds to the original decision problem cost, and in all but a few cases the feasible space is fixed, i.e. $C(x, y) = C$.
The connection between $y$ and $x_+$ implies that ground truth information is available for the ML model. This fact is exploited in DFL to craft customized loss functions, which can improve convergence and solution quality for the training process.

Our approach is considerably more flexible than DFL, having the ability to deal with general constraints and cost functions, and sequential decision-making. However, \ouracronym{} does not strictly generalize DFL, due to our use of the true cost $f$ as an objective at training time. This fact also implies that \ouracronym{} might be subject to the same difficulties as some early DFL approaches (e.g. convergence issues when $y$ refers to linear cost coefficients). That said, 1) in those cases, using DFL might be more appropriate; 2) it should be possible to adapt DFL losses in \ouracronym{}, and conversely some of our ideas (e.g. virtual parameters) could be adapted to DFL.

% CONSTRAINED RL
\paragraph{Constrained RL}

A variety of approaches have been investigated to incorporate constraints in (Deep) Reinforcement Learning, and a good survey on the topic is provided by \cite{liu2021policy}.

\emph{A first class of techniques} attempts to learn within the same ML model how to satisfy the constraints and how to maximize the reward.
One way this is done is via reward shaping \cite{grzes2017reward}: constraint violation is modeled as a negative reward, similarly to what is done in the classical penalty method. Just as for the penalty method, using a large penalty is effective at enforcing constraint satisfaction, but can cause numerical instabilities and lead to poorly optimized policies. As a result, calibrating the penalty value is a non-trivial task.
Another strategy, investigated by \cite{yang2019projection}, enforces constraints by projecting the policy parameters in the pre-image of the feasible space. In other words, after a gradient update, a projection step adjusts the weights so that the decision vector becomes feasible, as in the projected gradient method \cite{parikh2014proximal}. This approach is more numerically stable, but also more computationally expensive.

Since all methods mentioned so far attempt to learn constraint satisfaction by changing the policy parameters, infeasible decisions may still occur on unseen examples due to generalization errors. For this reason, \emph{a second class of methods} enforces constraint satisfaction via a projection step in decision space, rather than in weight space. In other words, once a baseline policy has provided a decision vector, this is projected in feasible space by minimizing a Euclidean distance. This step is often presented as a safety layer on top of a Neural Network policy \cite{dalal2018safe}. This technique can guarantee constraint satisfaction, and it is more scalable than projection in weight space. However, projection can still lead to lower rewards.

The first class of approaches can be seen as approximately solving \Cref{eqn:raw_training}. The second class can be assimilated to \Cref{eqn:om_seq_training}, except that the ML output is already a decision vector (i.e. $y \in Z$), and that $\tilde{f}(x, y, z)$ is always the Euclidean distance $\|y - z\|_2$. Our method therefore generalizes both classes of approaches. As the main appeal, the use of an arbitrary optimization problem in \ouracronym{} frees the ML model from the need to output a decision vector, and opens up the possibility to solve a difficult task by partitioning its complexity between the model $h$ and the function $g$.

% ALGORITHM CONFIGURATION
\paragraph{Algorithm Configuration}

Many approaches in the fields of Algorithm Configuration, Parameter Tuning, and Auto ML are concerned with adjusting the parameters of a given algorithm so as to optimize its behavior on a target distribution. A good survey is provided by \cite{schede2022survey}.

For example, the approaches in \cite{hutter2011sequential,hutter2014algorithm} consist of surrogate-based black-box optimization procedures to select the parameters of a target algorithm, so as to maximize its performance on a given benchmark. A similar, but more rigorous, result is achieved in \cite{DBLP:journals/jair/FilippoLM21} for myopic online algorithms; the proposed technique is however only applicable to convex problems.
Instance-specific configuration algorithms, such as the one from \cite{kadioglu2010isac,xu2008satzilla}, have the same goal, but they are capable of selecting different parameter values depending on the properties of the input instance.

The Dynamic Algorithm Configuration (DAC) approach from \cite{biedenkapp2020dynamic} is capable of adjusting the parameter values as the target algorithm progresses, which can provide advantages for iterative optimization processes (e.g. gradient descent) and sequential decision making. DAC works by casting the configuration problem as a Contextual MDP and using RL to obtain a policy.

Traditional Algorithm Configuration methods can be interpreted as solving \Cref{eqn:om_training}, while DAC as solving the sequential formulation from \Cref{eqn:om_seq_training}.
In all cases but one, the vector $y$ corresponds to algorithm parameters (e.g. learning rates), rather than model parameters.
Moreover, these approaches treat the algorithm to be configured as fixed, and therefore they do not take advantage of the flexibility offered by virtual parameters. To the best of our knowledge, the only applications of Algorithm Configuration together with true virtual parameters are provided by \cite{dickerson2012dynamic} for the Kidney Exchange Problem and \cite{silvestri2022hybrid} for an Energy Management System use case; however, these approaches have never been generalized.

% STOCHASTIC OPTIMIZATION
\paragraph{Stochastic Optimization}

The area of Stochastic Optimization \cite{powell2019unified} also aims at improving robustness in single- or multi-stage decision-making problems.

Most stochastic optimization algorithms rely on Monte Carlo methods to approximate expected values and assess constraint satisfaction. The Sample Average Approximation (SAA) has long been a staple of the field; its convergence rate in the context of combinatorial problems was eventually provided in \cite{kleywegt2002sample}. The SAA has been combined with Benders decomposition to address two-stage decision problems, leading to the family of L-shaped methods for stochastic optimization \cite{van1969shaped,laporte1993integer}. Multi-stage (i.e. sequential) decision problems have been often approximated as a sequence of 2-stage problems in so-called online anticipatory algorithms \cite{hentenryck2006online}, and solved once again via L-shaped methods. The connection between online stochastic optimization and Markov Decision Processes has been instead exploited directly by the AMSAA method from \cite{mercier2008amsaa}.

The stochastic optimization algorithms discussed here can be interpreted as addressing \Cref{eqn:om_training}, or \Cref{eqn:om_seq_training} in the case of AMSAA, without making use of a ML model, i.e. with $h(x, \theta) = x$. Because of that, there is no training step and solving a new instance requires approximating the expected value from scratch, making the process quite expensive. Our formulation suggests how by introducing a ML model the computational cost of sampling can be paid in a single offline phase, making inference much more efficient.

% EMPIRICAL EVALUATION
\section{Empirical Evaluation}%
\label{sec:experiments}

Our experimental evaluation is designed to show how \ouracronym{} can be used to achieve different results, and how a decision problem can be refactored, assigning its parts to the most suitable component among $h$ or $g$.
In particular, we use our approach to: 1) perform tuning of virtual parameters, similarly to Algorithm Configuration; 2) enforce constraint satisfaction in Reinforcement Learning; 3) obtain results analogous to DFL in a scenario where existing approaches are not applicable; 4) improve decision robustness without relying on the Sample Average Approximation.
%
    % \begin{itemize}
    %     \item \textbf{Constraints in Reinforcement Learning}. Learning to satisfy constraints is a hard task for a Reinforcement Learning algorithm whereas they can be easily managed by constrained optimization methods.
    %     \item \textbf{Generalization of Decision-Focused Learning.} We will show a practical use case where adopting \ouracronym{} is more straightforward than Decision Focused Learning approaches.
    %     \item \textbf{Low price solutions for stochastic optimization problems.} A \ouracronym{} implementation for solving stochastic optimization problems under uncertainty can provide good solutions at a reasonable computational cost compared to a Sampling Average Approximation method. 
    % \end{itemize}

% MODEL PARAMETER TUNING
\subsection{Model Parameter Tuning}

In this section, we approach model parameter tuning as an instance of \ouracronym{} exploiting its properties, namely adding virtual parameters to the optimization model and framing the task as a sequential decision-making problem.

The EMS is a good candidate for this experiment since it has a straightforward sequential formulation. Moreover, the online optimization problem employed at each stage is suboptimal because it does not take into account the full-time horizon.
Since the online optimal solver is myopic and suboptimal, we can improve its behavior by introducing a virtual cost parameter. Storing energy has no profit so the online (myopic) solver always ends up selling all the energy on the market. In particular, for this problem, we can define a virtual cost parameter related to the storage system (referred to as $c_{st}$), and it is possible to associate a profit (negative cost) with storing energy, which enables addressing this limitation. Intuitively, it is desirable to encourage the online solver to store energy in the battery system when the prices of the Electricity Market are cheap and the loads are low, in anticipation of future higher users' demand.

\begin{figure}[h]
    \centering
    \includegraphics[width=0.45\textwidth]{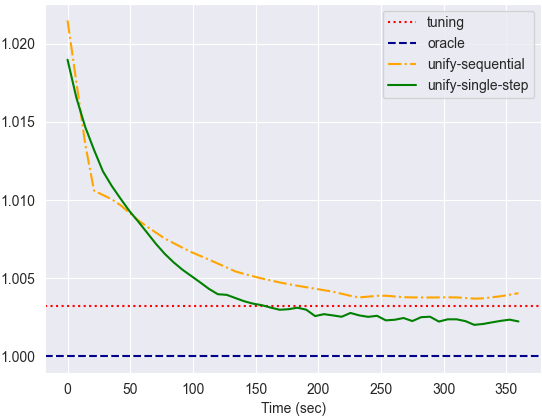}
    \caption{Optimality gap of the state-of-the-art \textsc{tuning} approach and the \ouracronym{} methods w.r.t. the computational time. }
    \label{fig:modelparamtuning}
\end{figure}

In \cite{def,de2021integrated}, the authors develop \textsc{tuning}, a hybrid offline/online optimization algorithm to find the optimal model parameters, but they assume a convex online optimization problem. Recently, \cite{silvestri2022hybrid} have explored the idea of using learning-based approximations as black-box solvers to perform model parameter tuning in a simpler integration scheme. In particular, their method employs Deep Reinforcement Learning as a black-box tool to find the optimal $c^{st}$ of the EMS described above.

In the following we will show that both the \textsc{single-step} and \textsc{mdp} methods proposed by \cite{silvestri2022hybrid} are instances of \ouracronym{}.
The former optimizes the parameters all at once, similarly to Algorithm Configuration approaches: according to \Cref{{eqn:om_training}}, the ML model $h$ provides the virtual costs for all the stages, namely $y = c^{st}_{\{1, \dots 96\}}$. Conversely, \textsc{mdp} exploits the sequential nature of the problem and, accordingly to \Cref{eqn:om_seq_training}, the model output at each step $k$ is $y_k = c^{st}_k$. We refer to them respectively as \ouracronym{}\textsc{-single-step} and \ouracronym{}\textsc{-sequential}. 
% While instantiating a \textsc{pro} approach is straightforward for this use case, using supervised Decision-Focused Learning approaches is not viable because one would require ground truth about the virtual cost.

Similarly to \cite{silvestri2022hybrid}, we compare the methods ensuring they have access to the same computation time. Since the time execution of \textsc{tuning} is constant, we chose this value as time limit for the training of the other methods and plot it as a horizontal line.
In \Cref{fig:modelparamtuning}, we show the optimality gap w.r.t. the computation time. As also highlighted in \cite{silvestri2022hybrid}, exploiting the sequential nature of the problem does not provide clear benefits (possibly due a suboptimal training solution) and \ouracronym{}\textsc{-single-step} overcomes both \textsc{tuning} and \ouracronym{}\textsc{-sequential}.

% CONSTRAINTS IN RL
\subsection{Constraints in RL}

Deep Reinforcement Learning DRL) approaches have trouble dealing with combinatorial structures and hard constraints in general. Our \ouracronym{} formulation provides a solution by delegating constraints satisfaction (and decision space exploration) to a second solver that can handle them more effectively. By doing so, the DRL algorithm can focus on a simpler task. 
%: this issue could be addressed by injecting knowledge of the online solver into the policy itself. We show that the resulting hybrid approach would benefit from powerful learning algorithms, and be well suited to deal with operational constraints.

%
\begin{figure}[h]
    \centering
    \includegraphics[width=0.45\textwidth]{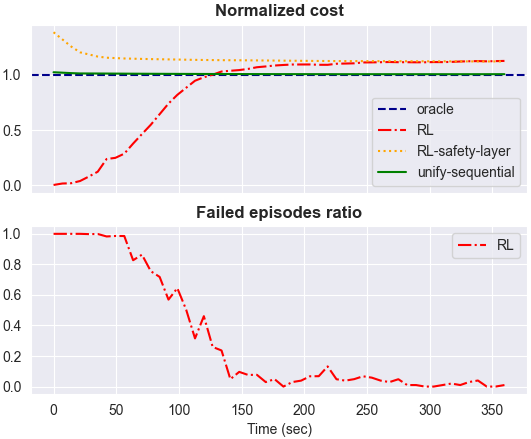}
    \caption{In this figure we show how demanding constraints satisfaction to the downstream solver greatly improves a full end-to-end RL method and \textsc{safety-layer}}.
    \label{fig:constrainedrl}
\end{figure}

For this experiment we use again the EMS use case as a benchmark. The problem is challenging because a policy must satisfy hard instantaneous constraints at each stage, i.e. the flow bounds and the power balance constraints. We train a full end-to-end DRL algorithm to provide a solution, by learning constraints satisfaction only from the reward signal and refer to it as \textsc{rl}. The design of the reward function is not trivial because it should provide a good trade-off between finding good solutions and exploring the feasible space. Projection-based DRL algorithms (e.g. Safety Layer) provide an alternative to full end-to-end methods when dealing with constraints: we have experimented with a Safety Layer implementation for the EMS and we will refer to it as \textsc{safety-layer.}

As shown in \Cref{sec:generalization}, both \textsc{safety-layer} and \textsc{\ouracronym{}-sequential} are instances of our \ouracronym{} framework, but they have a critical difference: in \textsc{safety-layer} the projection step can fix infeasible decisions, but this is done in a cost-agnostic fashion and the RL agent still needs to output a meaningful decision vector; conversely, in \textsc{\ouracronym{}-sequential} the RL agent is indirectly guiding a problem-specific solver, which has the same guarantees in terms of feasibility, but it is arguably a simpler task.

In the upper and lower parts of \cref{fig:constrainedrl}, we respectively show the optimality gap of all the methods and the number of failed episodes of \textsc{rl} due to constraints violations. In the early stages of training, \textsc{rl} never completes a full episode. It then progressively learns to satisfy constraints but, conversely, the cost of the solutions found increases. On the other hand, \textsc{safety-layer} converges very quickly but the final solution cost is very close to the one provided by \textsc{rl}. \ouracronym{}\textsc{-sequential} quickly converges as well, and it also \textit{improves the previous methods of a non-negligible gap.} 

These experimental results demonstrate that Reinforcement Learning can benefit from a policy reformulation that properly balances learning and optimization.

% GENERALIZATION OF DFL
\paragraph{Generalization of DFL}

In this section, we will show that \ouracronym{} can be used in place of classical Decision Focused Learning approaches.

For this experiment, we will consider the Set Multi-cover Problem with stochastic coverage requirements and make a comparison with a traditional predict-then-optimize approach.
In the \ouracronym{} formulation for this use case, the RL policy predicts the coverage requirements, $y=\{d_j\}$ $\forall j \in N$ that are then plugged into the optimization model, and the overall policy is trained to minimize the solution cost. The \ouracronym{} implementation is straightforward, whereas applying traditional Decision Focused Learning approaches is not trivial since the cost function is non-linear and non-differentiable.
In the predict-then-optimize approach, we first train a ML model to accurately predict the rates of the underlying Poisson distribution that generates the stochastic coverage requirements, and then we use the predicted rates as the requirements in the Set Multi-cover optimization model. 

\begin{figure}[h]
    \centering
    \includegraphics[width=0.45\textwidth]{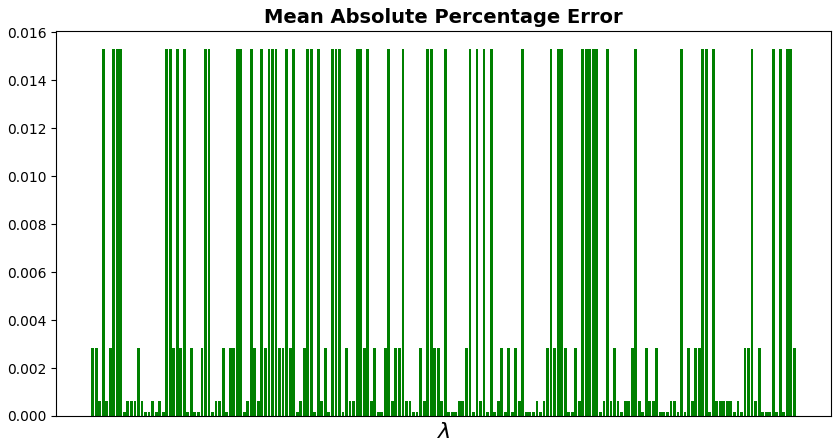}
    \caption{Here we report the predicted $\lambda$ MAPE of the Machine Learning model employed in the predict-then-optimize approach.}
    \label{fig:mape}
\end{figure}

\begin{figure}[h]
    \centering
    \includegraphics[width=0.45\textwidth]{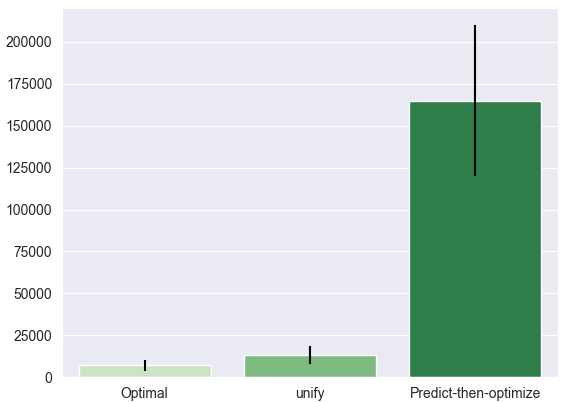}
    \caption{This figure reports the mean cost and standard deviation for the \ouracronym{} and the Predict-then-optimize approaches compared to the optimal values.}
    \label{fig:dflgeneralization}
\end{figure}

For training and testing the methods, we used two separated sets of instances. As we can see in \cref{fig:mape}, the Mean Absolute Percentage Error (MAPE) for each of the rate $\lambda_i$ $\forall i \in I$ is low, proving that the ML model is accurate. Despite this fact, as shown in \cref{fig:dflgeneralization}, the true task loss, i.e. the solution cost, is far from optimal. On the other side, the \ouracronym{} implementation that is trained to directly minimize the task loss greatly outperforms the predict-then-optimize approach and it is considerably closer to the optimal cost. We can thus conclude that \textit{\ouracronym{} is a valid alternative to traditional Decision Focused Learning approaches}, when those cannot be easily applied. 

%

% STOCHASTIC OPTIMIZATION
\subsection{Stochastic Optimization}

Solving stochastic optimization problems can be incredibly challenging. As mentioned in \cref{sec:generalization}, SAA methods are widely adopted in this field but they can be computationally expensive. In this section we will show how \ouracronym{} can be used to improve the robustness of the downstream solver by performing a set of experiments on the Set Multi-cover with stochastic coverages.

As a baseline approach, we implemented a SAA method that computes the optimal solution on a fixed set of instances (referred to as training instances). As also described in \cref{sec:generalization}, this approach is a very simple instance of \ouracronym{} without the Machine Learning component: it basically ignores the relationship between the observable variable $x$ and the Poisson rates $\lambda$.
We also provide a more robust baseline method that relies on the predict-then-optimize framework: we train a Machine Learning model on the training instances to estimate the parametrization of the probability distribution that models the uncertain variables; we then exploit the model predictions to generate samples for the downstream optimization model. Finally, we also train a more advanced implementation of \ouracronym{}, where the RL policy predicts the demands and the overall policy is trained to minimize the task loss, the same as described in the previous paragraph; notably, the latter approach does not rely on sampling. It is worth highlighting that the comparison favors the SAA method, since it is designed by assuming exact knowledge of the type of probability distribution; the \ouracronym{} implementation makes no such assumption.

\begin{figure}[h]
    \centering
    \includegraphics[width=0.45\textwidth]{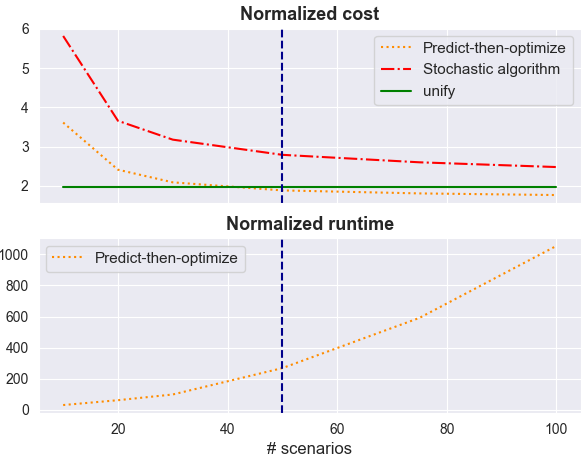}
    \caption{Optimality gap of the three methods on the Set Multi-cover problem with stochastic demands and the solution time of the predict-then-optimize approach w.r.t. the number of scenarios.}
    \label{fig:replacingsampling}
\end{figure}

Results are shown in \cref{fig:replacingsampling}. In the upper part of the figure, we report the optimality gap of the three methods on a separated set of instances w.r.t. the number of sampled scenarios. Both the simple SAA algorithm and predict-then-optimize approaches benefit from increasing the number of sampled scenarios.
On the other side, the \ouracronym{} implementation does not depend on the number of sampled scenarios, because it directly predicts the coverage values that are then plugged into the optimization model. Despite the advantage previously discussed, the predict-then-optimize approach surpasses \ouracronym{} only when at least $\sim 50$ scenarios are used in the downstream stochastic optimization model. 
In the lower part of the figure, we show the runtime required by predict-then-optimize as a multiple of the runtime of \ouracronym{}, w.r.t. to the number of scenarios. As we can see, to obtain better results, predict-then-optimize requires more than $200$ times the computation of \ouracronym{}.

We can thus conclude that a \textit{smart implementation of \ouracronym{} is a cheaper alternative to a SAA method for improving the robustness of the solver when facing stochastic optimization problems.} 

% CONCLUDING REMARKS
\section{Concluding Remarks}%
\label{sec:conclusion}

In this paper we proposed \ouracronym{}, a flexible framework for solving CO problems with ML. At the price of a higher design effort, we can face complex problems more easily by reformulating the original policy in two components: a ML model and a traditional constrained optimization problem.
We showed that \ouracronym{} provides a unified view of several approaches, which we hope will improve cross-fertilization by highlighting previously unrecognized connections.
Finally, we performed an extensive experimental evaluation of two practical problems to highlight the benefits of the approach.

Several open questions remain and we hope they will encourage future works in this direction. As mentioned in \cref{sec:generalization}, \ouracronym{} is still not a complete generalization of DFL and we should find a way to adapt surrogate losses defined by DFL methods in \ouracronym{}. In addition, Reinforcement Learning has been recently employed to build a solution for combinatorial optimization problems or inside the search process of already existing solvers. With some effort, it may be possible to reformulate these methods in the \ouracronym{} framework.

Despite the degree of success obtained in recent years, Reinforcement Learning has still some issues that limit its applicability on real-world problems. To name one of them, Reinforcement Learning is sample inefficient \cite{yu2018towards}: it requires several interactions with the environment before reaching convergence. It could be interesting to investigate whether improving the communication of the two \ouracronym{} components can lead to faster convergence at training time.

Problems where the decision space changes depending on the observable variables or where constraints are defined over multiple steps are challenging for traditional DFL approaches. As part of future works, we would like to study how \ouracronym{} can be used to face this kind of problem.

\clearpage

% \begin{itemize}
%     \item Adapt DFL losses for \ouracronym{}
%     \item Test a use case where the decision space size changes depending on the observables
%     \item Can the use of a CO component improve sample efficiency in RL, or improve convergence? This might require to improve communications between the components
%     \item Use RL to relax some constraints (e.g. chance constraints)
%     \item Constraints over cumulative quantities defined over multiple steps (those could be handle in the ML components)
%     \item Generalize approaches that use RL to construct or search for a solution
% \end{itemize}

% Use \bibliography{yourbibfile} instead or the References section will not appear in your paper

% APPENDIX

\appendix

% SET MULTI-COVER
\section{Set Multi-cover with stochastic coverage requirements}
\label{sec:smc}

We start from the formulation of the deterministic version of the problem:
\begin{align}
        \min & \sum_{j \in J} c_j x_j\\
        \sum_{j \in J} & a_{i, j} x_j \geq d_i \quad \forall i \in I\\
        x_j & \geq 0\\
        x_j & \in \mathbb{Z}\\
        a_{i,j} & \in \{ 0,1 \}
    \end{align}
where $I$ is the universe, $J$ is the collection of all possible sets, $a$ is the availability matrix, $d_i$ is the required coverage for the $i$-th element of the universe, $c_j$ is the cost of the $j$-th set and $x$ is the vector of decision variables.

In the stochastic version of the problem, the coverage requirements are sampled from a Poisson distribution with rate $\lambda_i$ $\forall i \in I$, namely $d_i \sim Poisson(\lambda_i)$.

% STOCHASTIC OPTIMIZATION MODEL
\paragraph{Stochastic optimization model}

As a baseline method to solve the Set Multi-cover with stochastic coverages, we implemented a Sample Average Approximation (SAA) algorithm based on Monte Carlo sampling. 

The optimization model is defined in the following way:
    \begin{align}
        \min \sum_{j \in J} c_j x_j & + \frac{1}{|\Omega|} \sum_{\omega \in \Omega} \sum_{i \in I} w_{i, \omega} s_{i, \omega} & \label{stoalg:costfun} \\
        \sum_{j \in J} a_{i, j} x_j & \geq d_{i, \omega} (1 - z_{i, \omega}) & \label{stoalg:demands} \\
        & & \forall i \in I, \omega \in \Omega \nonumber \\ 
        z_{i, \omega} = 1 & \implies s_{i, \omega}  \geq d_{i, \omega} - \sum_{j \in J} a_{i, j} x_j & \label{stoalg:indicatorconstr} \\
        & &\forall i \in I, \omega \in \Omega \nonumber \\
        x_j & \geq 0 \\
        z_{i, \omega} & \in \left[ 0, 1 \right] \\
        s_{i, \omega} & \geq 0 \\
        x, z & \in \mathbb{Z}
    \end{align}
where $\omega \in \Omega$ are the sampled scenarios, $w$ is the penalty vector due to the violation of the coverage requirements constraints of \cref{stoalg:demands}, $z$ is a vector of indicator variables that allow violation of \cref{stoalg:demands} and $s$ is a vector of slack variables that keep track of the not satisfied demands.

% ENERGY MANAGEMENT CASE STUDY
\paragraph{Energy Management System}
\label{sec:EMS}

An Energy Management System (EMS) requires the allocation of the minimum-cost power flows from different Distributed Energy Resources (DERs). The uncertainty stems from uncontrollable deviations from the planned loads of consumption and the presence of Renewable Energy Sources (RES). 
Based on actual energy prices and on the availability of DERs, the EMS decides: 1) how much energy should be produced; 2) which generators should be used for the required energy; 3) whether the surplus energy should be stored or sold to the energy market. 
Unlike in most of the existing literature, we acknowledge that in many practical cases \cite{de2020blind} \emph{some parameters can be tuned offline}, while the energy balance should be maintained online by managing energy flows among the grid, the renewable and traditional generators, and the storage systems. Intuitively, handling these two phases in an integrated fashion should lead to some benefits, thus making the EMS a good benchmark for our integrated approach. 

In our case study, it is desirable to encourage the online heuristic to store energy in the battery system when the prices of the Electricity Market are cheap and the loads are low, in anticipation of future higher users' demand. Storing energy has no profit so the online (myopic) solver always ends up selling all the energy on the market. However, by defining a \emph{virtual cost} parameter related to the storage system, it is possible to associate a profit (negative cost) with storing energy, which enables addressing this greedy limitation. Then, based on day-ahead RES generation and electric demand forecasts, we can find the optimal virtual costs related to the storage system to achieve better results in terms of solution quality (management costs of the energy system).

% STATE-OF-THE-ART OFFLINE/ONLINE APPROACH
\section{State-of-the-art Offline/Online approach}
\label{sec:sotamethod}

We refer to \textsc{tuning} as the integrated offline/online optimization method proposed in \cite{de2020blind,de2021integrated} that assumes \emph{exogenous uncertainty}, and that is composed of two macro steps: an offline two-stage stochastic optimization model based on sampling and scenarios; and an online parametric algorithm, implemented within a simulator, that tries to make optimal online choices, by building over the offline decisions. The authors assume that \emph{the online parametric algorithm is based on a convex optimization model}. Based on some configuration parameters of the online model, an offline parameter tuning step is applied. In this way, the authors can take advantage of the convexity of the online problem to obtain guaranteed optimal parameters. In particular, convexity implies that any local minimum must be a global minimum. Local minima can be characterized in terms of the KKT optimality conditions. Essentially, \emph{those conditions introduce a set of constraints that must be satisfied by any solution that is compatible with the behavior of the online heuristic}. They can exploit this property by formulating the tuning phase as a Mathematical Program that is not a trivial task for every constrained real-world problem. 

The online step is composed by a greedy (myopic) heuristic that minimizes the cost and covers the energy demand by manipulating the flows between the energy sources. We underline that this is a typical approach to handle the online optimization of an EMS \cite{6162768}. The heuristic can be formulated as an LP model:

% BASELINE HEURISTIC
\begin{align}
    \min & \sum_{k=1}^n \sum_{g \in G} c_g^k x_g^k \label{eqn:onlineobj} \\
    \text{s.t.} \ & \Tilde{L}^k = \sum_{g \in G} x_g^k \label{eqn:onlinepowerbalance} \\
    & 0 \leq \gamma_k + \eta x_0^k \leq \Gamma \label{eqn:onlinestorage} \\
    & \underbar{x}_g \leq x_g^k \leq \overline{x}_g \label{eqn:onlinebounds}
\end{align}
For each stage $k$ up to $n$, the decision variables $x_g$ are the power flows between nodes in $g \in G$ and $c_g$ are the associated costs. All flows must satisfy the lower and upper physical bounds $\underbar{x}_g$ and $\overline{x}_g$. Index $0$ refers to the storage system and the index $1$ to the RES generators. Hence the virtual costs associated with the storage system are $c_0^k$. The battery charge, upper limit and efficiency are $\gamma$, $\Gamma$ and $\eta$. The EMS must satisfy the user demand at each stage $k$ referred to as $\Tilde{L}^k$.

The baseline offline problem is modeled via MIP and relies on the KKT conditions to define a model for finding the optimal values of $c_0^k$ for the set of sampled scenarios $\omega \in \Omega$. Such model is given by:
%
% OFFLINE PROBLEM
\begin{align}
    \min & \frac{1}{|\Omega|} \sum_{\omega \in \Omega} \sum_{g \in G} \sum_{k=1}^n c_g^k x_{g, \omega}^k \label{eqn:objfunc} \\
    \text{s.t.} \ & \Tilde{L}_\omega^k = \sum_{g \in G} x_{g, \omega}^k & \forall{\omega} \in \Omega, \forall{k}=1, \cdots, n \label{eqn:powerbalance} \\
    & \underbar{x}_g \leq x_{g, \omega}^k \leq \overline{x}_g & \forall{\omega} \in \Omega, \forall{k}=1, \cdots, n \label{eqn:flowsranges} \\
    & 0 \leq \gamma_{\omega}^{k} \leq \Gamma & \forall{k}=1, \cdots, n \label{eqn:batteryrange} \\
    & \gamma_{\omega}^{k+1} = \gamma_{\omega}^{k} + \eta x_{0, \omega}^k & \forall{\omega} \in \Omega, \forall{k}=1, \cdots, n-1 \label{eqn:batterytrans} \\
    & x_{1, \omega}^{k+1} = \hat{R}_k + \xi_{R, \omega}^k & \forall{\omega} \in \Omega, \forall{k}=1, \cdots, n \label{eqn:restrans} \\
    & \Tilde{L}_{\omega}^{k+1} = \hat{L}_k + y_k + \xi_{L, \omega}^k & \forall{\omega} \in \Omega, \forall{k}=1, \cdots, n 
    \label{eqn:loadtrans} 
\end{align}
$\hat{R}_k$ and $\hat{L}_k$ are the estimated RES production and load, and $\xi_{R}^k$ and $\xi_L^k$ are the corresponding random variables representing the prediction errors. $y_k$ are optimal load shifts and are considered as fixed parameters. 
The authors assume that the errors follow roughly a Normal distribution $N(0, \sigma^2)$ and that the variance $\sigma^2$ is such that $95\%$ confidence interval corresponds to $\pm10\%$ of the estimated value. $\Tilde{L}^k_{\omega}$ is the observed user load demand for stage $k$ of the scenario $\omega$.
\Cref{eqn:batterytrans,eqn:restrans,eqn:loadtrans} model the transition functions.

The above formulation is free to assign variables (as long as the constraints are satisfied), whereas all decisions that are supposed to be made by the heuristic can not rely on future information. The authors account for this limitation by introducing, as constraints, the KKT optimality conditions for the convex online heuristic. The model achieves integration at the cost of offline computation time, because of the additional variables introduced and the presence of non-linearities.

In the following we show the KKT conditions formulation for the online heuristic in a single scenario:
\begin{align}
  & -c_{g}^k = \lambda_{\omega}^k + \mu_{g,\omega}^{k} - \nu_{g,\omega}^k & \forall g \in G \label{eqn:vppkkt1}\\
  & \mu_{g,\omega}^k (x_{g,\omega}^{k} + \overline{x}_{g}) = 0 & \forall g \in G\\
  & \nu_{i,\omega}^{k}(\underline{x}_{g} - x_{g,\omega}^{t}) = 0 & \forall g \in G \\
  & \hat{\mu}_{\omega}^k (\eta x_{0,\omega}^{k} + \gamma^k - \Gamma) = 0 \\
  & \hat{\nu}_{\omega}^k (\eta x_{0,\omega}^{k} + \gamma^k) = 0 \\
  & \mu_{g,\omega}^{k}, \nu_{g,\omega}^{k} \ge 0 & \forall g \in G \\
  & \hat{\mu}_{\omega}^{k}, \hat{\nu}_{\omega}^{k} \ge 0 \label{eqn:vppkkt2}
\end{align}
where $\mu_{g,\omega}^k$ and $\nu_{g,\omega}^k$ are the multipliers associated to the physical flow bounds, while $\hat{\mu}_{\omega}^k$ and $\hat{\nu}_{\omega}^k$ are associated to the battery capacity bounds. Injecting the conditions in the offline model yields:
\begin{align}
  \min\ & \frac{1}{|\Omega|}\sum_{\omega \in \Omega} \sum_{g \in G} \sum_{k = 1}^n c_{g}^k x_{g,\omega}^{k} \hspace{10mm} \nonumber \\
\text{s.t. }
& \textrm{Eq. } ~\eqref{eqn:powerbalance}-\eqref{eqn:loadtrans} \hspace{47mm}  \quad \nonumber \\
& \textrm{Eq. } ~\eqref{eqn:vppkkt1}-\eqref{eqn:vppkkt2} \hspace{10mm} \forall \omega \in \Omega, \forall k = 1,\ldots n \hspace{2mm}  \quad \nonumber
\end{align}

where the decision variables are $x_{g,\omega}^k$, $\mu_{g,\omega}^{k}$, $\nu_{g,\omega}^{k}$, $\hat{\mu}_{\omega}^{k}$, $\hat{\nu}_{\omega}^{k}$. To those, the authors add the cost $c_0^k$ associated with the flow from and to the storage system (the only  parameter they allow the solver to adjust). This method allows the offline solver to associate a virtual profit for storing energy, which enables addressing the original limitation at no online computational cost.

% DATASETS
\section{Datasets}

In this section we describe the datasets used to conduct our experimental evaluation.

% ENERGY MANAGEMENT SYSTEM
\paragraph{Energy Management System}
For the EMS we used the same dataset of \cite{silvestri2022hybrid} that relies on real data based on a Public Dataset \footnote{www.enwl.co.uk/lvns}. From this dataset, we assume electric load demand and photovoltaic production forecasts, upper and lower limits for generating units and the initial status of storage units. The realizations of uncertain variables, namely $\tilde{R}$ and $\tilde{L}$, are obtained from the forecasts by adding noise from a normal distribution as described in \cref{sec:sotamethod}.
The dataset presents individual profiles of load demand with a time step of 5 minutes resolution from 00:00 to 23:00. We consider aggregated profiles with a timestamp of 15 minutes and use them as forecasted load. 
The photovoltaic production is based on the same dataset with profiles for different sizes of photovoltaic units but the same solar irradiance (i.e. the same shape but different amplitude due to the different sizes of the panels used). Also in this case photovoltaic production is adopted as a forecast.

The electricity demand hourly prices have been obtained based on data from the Italian national energy market management corporation\footnote{http://www.mercatoelettrico.org/En/Default.aspx} (GME) in €/MWh. 
The diesel price is taken from the Italian Ministry of Economic Development\footnote{http://dgsaie.mise.gov.it/} and is assumed as a constant for all the time horizon (one day in our model) as assumed in literature \cite{6162768} and from \cite{ewl}.

For the evaluation, we randomly select 100 pairs of load demand and photovoltaic production forecasts.

% SET MULTI-COVER WITH STOCHASTIC COVERAGES
\paragraph{Set Multi-cover with stochastic coverages}

For the Set Multi-cover with stochastic coverages, we generate the availability matrix $a$ following the guidelines of \cite{grossman1997computational}: every column covers at least one row and every row is covered by at least two columns. In addition, the availability matrix has a density (number of 1 in the matrix) of $2\%$. The set costs are randomly generated in the range $\left[ 1, 100 \right]$ with a uniform probability distribution. The penalties $w$ introduced in \cref{sec:smc} are computed as follows:
    \begin{equation}
        w_{i, \omega} = \max_{\substack{j \in J \\ a_{i,j} = 1 }}{c_j} \cdot 10 \quad \forall{\omega \in \Omega} \nonumber
    \end{equation}
The equation above means that the penalty for the $i$-th product is computed as the maximum set cost among the ones that cover the $i$-th element, multiplied by $10$. This basically ensures that covering an element is always more convenient than receiving a penalty. In addition, penalties are the same for all the scenarios.

We generate $1000$ instances with $200$ elements and $1000$ sets and we equally split them between training and test sets. The Poisson vector rates $\lambda$ is generated according to a linear relationship with an observable variable $o \in \mathbb{R}$, $\lambda = a_i o$ $\forall i \in I$. The coefficients $a$ and the observable variables vectors $o$ are randomly generated with an uniform probability distribution respectively in the range $\left[ 1, 5 \right]$ and $\left[ 1, 10 \right]$.

% REINFORCEMENT LEARNING ENVIRONMENTS
\section{Reinforcement Learning Environments}

In this section we provide a description of the environments design for the EMS and the Set Multi-cover with stochastic coverage requirements, namely observations, actions and reward function.

% ENERGY MANAGEMENT SYSTEM
\paragraph{Energy Management System}

In the EMS use case we employed the same environment variants introduced in \cite{silvestri2022hybrid}. For \textsc{\ouracronym{}-single-step} the observations are the day-ahead  photovoltaic generation $ \hat{R}^k $ and electric demand forecasting $ \hat{L}^k $ and the actions are the set of virtual costs for all the stages $c_0^k $ for $k = \{ 1, \dots, n\}$ and thus the policy is a function $\pi : \mathbb{R}^{n \times 2} \xrightarrow{} \mathbb{R}^n$.
Once $ c_0^k $ are provided, a solution $\{ x_g^k \}_{k=1}^n$ is found solving the online optimization problem defined in \Cref{eqn:onlineobj,eqn:onlinepowerbalance,eqn:onlinestorage,eqn:onlinebounds} and the reward is the negative real cost computed as:
\begin{align*}
   - \sum_{k=1}^n \sum_{ \substack{g \in G \\ g \neq 0 }} c_g^k x_g^k 
\end{align*}

For \textsc{\ouracronym{}-sequential} the policy $\pi$ is a function $\pi: \mathbb{R}^{ n \times 3 + 1 } \xrightarrow{} \mathbb{R}$. 
The state $s_k$ keeps track of the battery charge $\gamma^k$ and it is updated accordingly to the input and output storage flows. At each stage $k$, the agent's action $a_k$ is the virtual cost $c_0^k$ and the corresponding online optimization problem is solved. Then the environment provides as observations the battery charge $\gamma^k$, the set of forecasts $\tilde{R}_{1, \dots, n}$ and $\tilde{L}_{1, \dots, n}$, and a one-hot encoding of the stage $k$. The reward is again the negative real cost but for the only current stage $k$:
\begin{align*}
   - \sum_{ \substack{g \in G \\ g \neq 0 }} c_g^k x_g^k 
\end{align*}

In the full end-to-end Reinforcement Learning approach, the policy provides a $(|G|-1) $-dimensional vector corresponding to the power flows for a single stage. The actions are clipped in the range $[-1,1]$ and then rescaled in their feasible ranges $\left [\underline{x}_g, \overline{x}_g \right]$. Since one of the power flows has no upper bound $\overline{x}_g$, the authors have set its value so that the power balance constraint of \cref{eqn:powerbalance} is satisfied, reducing the actions space and making the task easier. We refer to this decision variable as $x_2^k$.
Despite adopting these architectural constraints, the actions provided by the agent may still be infeasible: the storage constraint of \cref{eqn:batterytrans} and the lower bound $\underline{x}_2$  can be violated. The reward is non-zero only for the last stage and it is computed as the negative cumulative real cost. Since the cost of the solution is in the range $\left[ 0, 3000 \right]$, the policy network is rewarded with a value of $-10000$ when infeasible actions are selected to encourage the search for feasible solutions. 

As last detail, for all the environments described above the observations are rescaled in the range $[0,1]$ dividing by their maximum values.

% SET MULTI-COVER WITH STOCHASTIC COVERAGES
\paragraph{Set Multi-cover with stochastic coverages}

For this use case, the environment has a single step duration: it provides the observable variable $o$ to the agent and the actions are the predicted coverages. The policy is thus a function $\pi: \mathbb{R} \xrightarrow{} \mathbb{Z}^I$ where $I$ is the set of the elements. Since the Reinforcement Learning agent outputs real numbers, we convert its actions to the closest integer values.
The reward is computed as the negative total cost:
    \begin{align}
        -\sum_{j \in J} & c_j \hat{x}_j - \sum_{i \in I} w_{i} \overline{d}_i \nonumber \\
        \overline{d}_i = & \max \left( 0, d_i - \sum_{j \in J} a_{i,j} x_j \right) \nonumber
    \end{align}
where $\hat{x}$ is the solution found using the predicted demands, $\overline{d}$ is the vector of not satisfied demands and $w$ is the vector of penalties as illustrated in \cref{sec:smc}.

% TRAINING AND HYPERPARAMETERS CONFIGURATION
\section{Training and hyperparameter configuration}

As Reinforcement Learning algorithm we employed the Advantage Actor-Critic (A2C) \footnote{A2C algorithm was implemented with the TensorFlow version of the \texttt{garage} \cite{garage} library.} because it is robust and can deal with continuous actions space. Since hyperparameter search was outside the scope of the paper, we chose a quite standard architecture. The policy is represented by a Gaussian distribution for each action dimension, parametrized by a feedforward fully-connected Neural Network with two hidden layers, each of 32 units and a hyperbolic tangent activation function. The critic is again a deep neural network with the same hidden architecture of the policy.

Parameters are updated using Adam optimizer. For the experiments on the EMS use case, we opted for a learning rate of $0.01$ (larger than usual) because it improved the convergence speed without compromising the final results.

For \textsc{\ouracronym{}-single-step} in the EMS use case and the \ouracronym{} implementation for the Set Multi-cover, we used a batch size of $100$ whereas for \textsc{\ouracronym{}-sequential}, \textsc{rl} and \textsc{safety-layer} we preferred a larger batch size of $9600$ to have a comparable number of episodes for each training epoch.

For the EMS experiments, \textsc{\ouracronym{}-single-step}, \textsc{\ouracronym{}-sequential}, \textsc{rl} and \textsc{safety-layer} have been trained for respectively $37, 19, 52$ and $19$ epochs. We chose these values because we want to provide the same computation time required by \textsc{tuning} algorithm. The \ouracronym{} implementation for the Set Multi-cover was trained for $10000$ epochs.

Experiments on the EMS were performed on a laptop with an Intel i7 CPU with 4 cores, 1.5 GHz clock frequency and 16 GB of memory. Experiments on the Set Multi-cover were conducted on a AMD EPYC 7272 16-Core Processor with 2.8 GHz clock frequency and 512 GB of memory. Despite the availability of multi-core processors, we do not exploit multi-threading and all experiments were executed on a single core to keep things the simplest as possible and simplify reproducibility.

\clearpage

\bibliographystyle{unsrt}  
\bibliography{references}

\begin{thebibliography}{10}

\bibitem{6162768}
D.~Aloini, E.~Crisostomi, M.~Raugi, and R.~Rizzo.
\newblock Optimal power scheduling in a virtual power plant.
\newblock In {\em 2011 2nd IEEE PES International Conference and Exhibition on
  Innovative Smart Grid Technologies}, pages 1--7, Dec 2011.

\bibitem{de2020blind}
Allegra {De Filippo}, Michele Lombardi, and Michela Milano.
\newblock The blind men and the elephant: Integrated offline/online
  optimization under uncertainty.
\newblock In {\em IJCAI}, 2020.

\bibitem{hua2009exact}
Qiang-Sheng Hua, Dongxiao Yu, Francis Lau, and Yuexuan Wang.
\newblock Exact algorithms for set multicover and multiset multicover problems.
\newblock In {\em International Symposium on Algorithms and Computation}, pages
  34--44. Springer, 2009.

\bibitem{donti2017task}
Priya Donti, Brandon Amos, and J~Zico Kolter.
\newblock Task-based end-to-end model learning in stochastic optimization.
\newblock {\em Advances in neural information processing systems}, 30, 2017.

\bibitem{DBLP:conf/ijcai/KotaryFHW21}
James Kotary, Ferdinando Fioretto, Pascal~Van Hentenryck, and Bryan Wilder.
\newblock End-to-end constrained optimization learning: {A} survey.
\newblock In Zhi{-}Hua Zhou, editor, {\em Proceedings of the Thirtieth
  International Joint Conference on Artificial Intelligence, {IJCAI} 2021,
  Virtual Event / Montreal, Canada, 19-27 August 2021}, pages 4475--4482.
  ijcai.org, 2021.

\bibitem{wilder2019melding}
Bryan Wilder, Bistra Dilkina, and Milind Tambe.
\newblock Melding the data-decisions pipeline: Decision-focused learning for
  combinatorial optimization.
\newblock In {\em Proceedings of the AAAI Conference on Artificial
  Intelligence}, volume~33, pages 1658--1665, 2019.

\bibitem{elmachtoub2017smart}
Adam~N Elmachtoub and Paul Grigas.
\newblock Smart “predict, then optimize”.
\newblock {\em Management Science}, 2021.

\bibitem{vlastelica2019differentiation}
Marin~Vlastelica {Pogančić}, Anselm {Paulus}, Vit {Musil}, Georg {Martius},
  and Michal {Rolinek}.
\newblock Differentiation of blackbox combinatorial solvers.
\newblock In {\em ICLR 2020 : Eighth International Conference on Learning
  Representations}, 2020.

\bibitem{DBLP:conf/nips/MandiG20}
Jayanta Mandi and Tias Guns.
\newblock Interior point solving for lp-based prediction+optimisation.
\newblock In Hugo Larochelle, Marc'Aurelio Ranzato, Raia Hadsell,
  Maria{-}Florina Balcan, and Hsuan{-}Tien Lin, editors, {\em Advances in
  Neural Information Processing Systems 33: Annual Conference on Neural
  Information Processing Systems 2020, NeurIPS 2020, December 6-12, 2020,
  virtual}, 2020.

\bibitem{DBLP:conf/ijcai/MulambaMD0BG21}
Maxime Mulamba, Jayanta Mandi, Michelangelo Diligenti, Michele Lombardi, Victor
  Bucarey, and Tias Guns.
\newblock Contrastive losses and solution caching for predict-and-optimize.
\newblock In Zhi{-}Hua Zhou, editor, {\em Proceedings of the Thirtieth
  International Joint Conference on Artificial Intelligence, {IJCAI} 2021,
  Virtual Event / Montreal, Canada, 19-27 August 2021}, pages 2833--2840.
  ijcai.org, 2021.

\bibitem{liu2021policy}
Yongshuai Liu, Avishai Halev, and Xin Liu.
\newblock Policy learning with constraints in model-free reinforcement
  learning: A survey.
\newblock In {\em IJCAI}, pages 4508--4515, 2021.

\bibitem{grzes2017reward}
Marek Grzes.
\newblock Reward shaping in episodic reinforcement learning.
\newblock 2017.

\bibitem{yang2019projection}
Tsung-Yen Yang, Justinian Rosca, Karthik Narasimhan, and Peter~J Ramadge.
\newblock Projection-based constrained policy optimization.
\newblock In {\em International Conference on Learning Representations}, 2019.

\bibitem{parikh2014proximal}
Neal Parikh, Stephen Boyd, et~al.
\newblock Proximal algorithms.
\newblock {\em Foundations and trends{\textregistered} in Optimization},
  1(3):127--239, 2014.

\bibitem{dalal2018safe}
Gal Dalal, Krishnamurthy Dvijotham, Matej Vecerik, Todd Hester, Cosmin
  Paduraru, and Yuval Tassa.
\newblock Safe exploration in continuous action spaces.
\newblock {\em arXiv preprint arXiv:1801.08757}, 2018.

\bibitem{schede2022survey}
Elias Schede, Jasmin Brandt, Alexander Tornede, Marcel Wever, Viktor Bengs,
  Eyke H{\"u}llermeier, and Kevin Tierney.
\newblock A survey of methods for automated algorithm configuration.
\newblock {\em arXiv preprint arXiv:2202.01651}, 2022.

\bibitem{hutter2011sequential}
Frank Hutter, Holger~H Hoos, and Kevin Leyton-Brown.
\newblock Sequential model-based optimization for general algorithm
  configuration.
\newblock In {\em International Conference on Learning and Intelligent
  Optimization}, pages 507--523. Springer, 2011.

\bibitem{hutter2014algorithm}
Frank Hutter, Lin Xu, Holger~H Hoos, and Kevin Leyton-Brown.
\newblock Algorithm runtime prediction: Methods \& evaluation.
\newblock {\em Artificial Intelligence}, 206:79--111, 2014.

\bibitem{DBLP:journals/jair/FilippoLM21}
Allegra {De Filippo}, Michele Lombardi, and Michela Milano.
\newblock Integrated offline and online decision making under uncertainty.
\newblock {\em J. Artif. Intell. Res.}, 70:77--117, 2021.

\bibitem{kadioglu2010isac}
Serdar Kadioglu, Yuri Malitsky, Meinolf Sellmann, and Kevin Tierney.
\newblock Isac--instance-specific algorithm configuration.
\newblock In {\em ECAI 2010}, pages 751--756. IOS Press, 2010.

\bibitem{xu2008satzilla}
Lin Xu, Frank Hutter, Holger~H Hoos, and Kevin Leyton-Brown.
\newblock Satzilla: portfolio-based algorithm selection for sat.
\newblock {\em Journal of artificial intelligence research}, 32:565--606, 2008.

\bibitem{biedenkapp2020dynamic}
Andr{\'e} Biedenkapp, H~Furkan Bozkurt, Theresa Eimer, Frank Hutter, and Marius
  Lindauer.
\newblock Dynamic algorithm configuration: foundation of a new meta-algorithmic
  framework.
\newblock In {\em ECAI 2020}, pages 427--434. IOS Press, 2020.

\bibitem{dickerson2012dynamic}
John~P Dickerson, Ariel~D Procaccia, and Tuomas Sandholm.
\newblock Dynamic matching via weighted myopia with application to kidney
  exchange.
\newblock In {\em Twenty-sixth AAAI conference on artificial intelligence},
  2012.

\bibitem{silvestri2022hybrid}
Mattia Silvestri, Allegra {De Filippo}, Federico Ruggeri, and Michele Lombardi.
\newblock Hybrid offline/online optimization for energy management via
  reinforcement learning.
\newblock In {\em International Conference on Integration of Constraint
  Programming, Artificial Intelligence, and Operations Research}, pages
  358--373. Springer, 2022.

\bibitem{powell2019unified}
Warren~B Powell.
\newblock A unified framework for stochastic optimization.
\newblock {\em European Journal of Operational Research}, 275(3):795--821,
  2019.

\bibitem{kleywegt2002sample}
Anton~J Kleywegt, Alexander Shapiro, and Tito Homem-de Mello.
\newblock The sample average approximation method for stochastic discrete
  optimization.
\newblock {\em SIAM Journal on Optimization}, 12(2):479--502, 2002.

\bibitem{van1969shaped}
Richard~M Van~Slyke and Roger Wets.
\newblock L-shaped linear programs with applications to optimal control and
  stochastic programming.
\newblock {\em SIAM journal on applied mathematics}, 17(4):638--663, 1969.

\bibitem{laporte1993integer}
Gilbert Laporte and Fran{\c{c}}ois~V Louveaux.
\newblock The integer l-shaped method for stochastic integer programs with
  complete recourse.
\newblock {\em Operations research letters}, 13(3):133--142, 1993.

\bibitem{hentenryck2006online}
Pascal~Van Hentenryck and Russell Bent.
\newblock {\em Online stochastic combinatorial optimization}.
\newblock The MIT Press, 2006.

\bibitem{mercier2008amsaa}
Luc Mercier and Pascal~Van Hentenryck.
\newblock Amsaa: A multistep anticipatory algorithm for online stochastic
  combinatorial optimization.
\newblock In {\em International Conference on Integration of Artificial
  Intelligence (AI) and Operations Research (OR) Techniques in Constraint
  Programming}, pages 173--187. Springer, 2008.

\bibitem{def}
Allegra {De Filippo}, Michele Lombardi, and Michela Milano.
\newblock Methods for off-line/on-line optimization under uncertainty.
\newblock In {\em IJCAI}, pages 1270--1276, 2018.

\bibitem{de2021integrated}
Allegra {De Filippo}, Michele Lombardi, and Michela Milano.
\newblock Integrated offline and online decision making under uncertainty.
\newblock {\em Journal of Artificial Intelligence Research}, 70:77--117, 2021.

\bibitem{yu2018towards}
Yang Yu.
\newblock Towards sample efficient reinforcement learning.
\newblock In {\em IJCAI}, pages 5739--5743, 2018.

\bibitem{ewl}
A.N. Espinosa and L.N. Ochoa.
\newblock Dissemination document “low voltage networks models and low carbon
  technology profiles”.
\newblock Technical report, University of Manchester, June 2015.

\bibitem{grossman1997computational}
Tal Grossman and Avishai Wool.
\newblock Computational experience with approximation algorithms for the set
  covering problem.
\newblock {\em European journal of operational research}, 101(1):81--92, 1997.

\bibitem{garage}
The garage contributors.
\newblock Garage: A toolkit for reproducible reinforcement learning research.
\newblock \url{https://github.com/rlworkgroup/garage}, 2019.

\end{thebibliography}

\end{document}